%% file: conference_101719.tex
\documentclass[conference]{IEEEtran}
\IEEEoverridecommandlockouts
\usepackage{cite}
\usepackage{amsmath,amssymb,amsfonts}
\usepackage{algorithmic}
\usepackage{graphicx}
\usepackage{textcomp}
\usepackage{xcolor}
\usepackage{our_imports}
\def\BibTeX{{\rm B\kern-.05em{\sc i\kern-.025em b}\kern-.08em
    T\kern-.1667em\lower.7ex\hbox{E}\kern-.125emX}}
\begin{document}

\title{OrthoNets: Orthogonal Channel Attention Networks}

\author{\IEEEauthorblockN{Hadi Salman}
\IEEEauthorblockA{\textit{Department of Computer Science and Computer Engineering} \\
\textit{University of Arkansas}\\
Fayetteville, Arkansas, USA \\
hs028@uark.edu}
\and
\IEEEauthorblockN{Caleb Parks}
\IEEEauthorblockA{\textit{Department of Computer Science and Computer Engineering} \\
\textit{University of Arkansas}\\
Fayetteville, Arkansas, USA \\
cgparks@uark.edu}
\and
\IEEEauthorblockN{Matthew Swan}
\IEEEauthorblockA{\textit{Department of Computer Science and Computer Engineering} \\
\textit{University of Arkansas}\\
Fayetteville, Arkansas, USA \\
ms163@uark.edu}
\and
\IEEEauthorblockN{John Gauch}
\IEEEauthorblockA{\textit{Department of Computer Science and Computer Engineering} \\
\textit{University of Arkansas}\\
Fayetteville, Arkansas, USA \\
jgauch@uark.edu}
}
\maketitle

\begin{abstract}
Designing an effective channel attention mechanism implores one to find a lossy-compression method allowing for optimal feature representation. Despite recent progress in the area, it remains an open problem.  FcaNet, the current state-of-the-art channel attention mechanism, attempted to find such an information-rich compression using Discrete Cosine Transforms (DCTs). One drawback of FcaNet is that there is no natural choice of the DCT frequencies. To circumvent this issue, FcaNet experimented on ImageNet to find optimal frequencies. We hypothesize that the choice of frequency plays only a supporting role and the primary driving force for the effectiveness of their attention filters is the orthogonality of the DCT kernels. To test this hypothesis, we construct an attention mechanism using randomly initialized orthogonal filters. Integrating this mechanism into ResNet, we create OrthoNet. We compare OrthoNet to FcaNet (and other attention mechanisms) on Birds, MS-COCO, and Places356 and show superior performance. On the ImageNet dataset, our method competes with or surpasses the current state-of-the-art. Our results imply that an optimal choice of filter is elusive and generalization can be achieved with a sufficiently large number of orthogonal filters. We further investigate other general principles for implementing channel attention, such as its position in the network and channel groupings. Our code is publicly available at (\url{https://github.com/hady1011/OrthoNets})
\end{abstract}


\section{Introduction}
\label{sec_intro}

Deep convolutional neural networks have become the standard tool to accomplish many computer vision tasks such as classification, segmentation, and object detection \cite{Faster_RCNN, Masked_RCNN}. Their success is attributed to the ability to extract features related to the underlying task. Higher quality features allow for better outputs in the decision space. Hence, improving the quality of these features has become an area of interest in the machine learning community \cite{AlexNet-V1, ResNeXt, FixResNext}. In this paper, we investigate channel attention mechanisms. 

A channel attention mechanism is a module placed throughout the network. Each module takes as input a feature, which has $C$ channels, and outputs a $C$ dimensional attention vector with components in $(0,1)$. By multiplying these outputs with the input feature vector, the features are adapted towards solving the underlying task.

\begin{figure}
    \centering
    \includegraphics[width=0.47\textwidth]{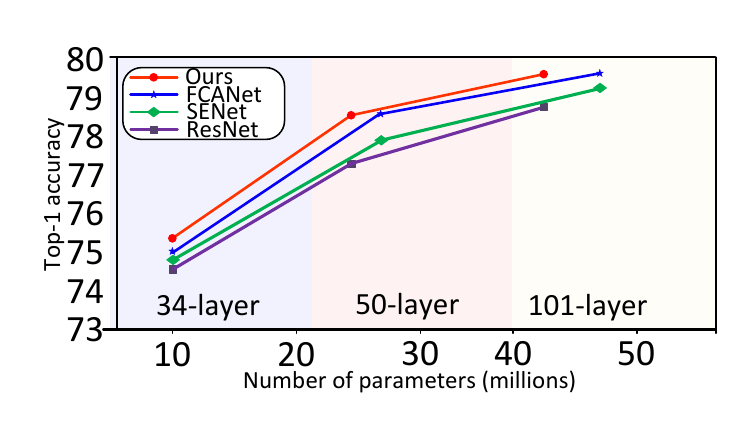}
    \label{fig:accuracies}
    \caption{ImageNet accuracy comparison. Our Method outperforms FcaNet using 34 layers and achieves competitive results with 50 and 101 layers while using 10\% less parameters and reducing computational cost.}
\end{figure}

The first channel attention mechanism, introduced by SENet \cite{SENet}, uses Global Average Pooling (GAP) to compress the spatial dimension of each feature channel into a single scalar. To compute the attention vector, the compressed representation is sent through a Multi-Layer Perceptron (MLP) and then through a sigmoid function. The compression stage is referred to as the squeeze. As pointed out in \cite{FcaNet}, a major weakness of SENet is using a single method (i.e.\ GAP) to compress each channel. This weakness was removed by FcaNet \cite{FcaNet}. They argue that GAP is discarding essential low-frequency information and that generalizing channel attention in the frequency domain by using DCTs, yields more information. Doing so improves channel attention significantly. Based on imperial results, FcaNet chose which Discrete Cosine Transforms (DCTs) to use, thus becoming the state-of-the-art (SOTA) channel attention mechanism. 

We hypothesize that the success of FcaNet has less to do with the low-frequency spectral information and is, in fact, due to the orthogonal nature of the DCT compression mappings. 
To test this hypothesis, we construct a channel attention mechanism using random orthonormal filters to compress the spatial information of each feature. 
Using our attention mechanisms, we produce networks that outperform FcaNet in object detection and image segmentation tasks. Compared to SOTA, our method achieves competitive or superior results on the ImageNet dataset and achieves top performance for attention mechanisms on Birds and Places365 datasets. We further investigate other general principles for implementing channel attention. The main contributions of this work are summarized as follows:

\begin{itemize}

    \item We propose to diversify the channel squeeze methods in an effort to extract more information using orthogonal filters and name the resulting network OrthoNet. By evaluating OrthoNet, we demonstrate that a key property of quality channel attention is filter diversity.
    
    \item We propose to relocate the channel attention module in ResNet-50 and 101 bottleneck blocks. Our location reduces the number of parameters used for our module and improves the network's overall performance.
    
    \item Running numerous experiments on Birds, Places365, MS-COCO, and ImageNet datasets; we demonstrate that OrthoNet is the state-of-the-art channel attention mechanism.
    
    \item To explore network architecture choices, we investigate channel grouping in the squeeze phase and squeeze filter learning.

\end{itemize}

The rest of the paper is formatted as follows. In section 2, we review works related to this paper. Section 3 formulates channel attention mechanism and reviews the most important channel attentions. In section 4, we report our results. Section 5 contains discussion on architecture choices and further benefits of this method. Finally, in section 6 we conclude our work and hypothesize why distinct attention filters are key to an attention mechanism. 

\section{Related Work}

\subsection{Deep Convolutional Neural Networks (DCNNs)} LeNet \cite{LeNet} and AlexNet \cite{AlexNet} served as the starting point for a new era in fast GPU-implementation of CNNs. Since then, researchers have started exploring the potential of adding more convolutional layers while maintaining computational efficiency. To improve the utilization of GPUs, GoogLeNet \cite{GoogLeNet} was proposed based on the Hebbian principle and the intuition of multi-scale processing. Shortly after, VGGNet \cite{VGGNet} was introduced and secured first and second place in ImageNet Challenge 2014. Their work began a trend to deepen networks to achieve higher accuracy. As a result, the number of network parameters increased causing difficulty during optimization. To overcome this challenge, ResNet \cite{ResNet} explicitly reformulated the layers as learning residual functions which made the network easier to optimize. Shortly after, many variants of ResNet were introduced to enhance it and overcome problems like vanishing gradients and parameter redundancy~\cite{WRN,ResNeXt,DenseNet}. 

%or redundant%
\subsection{Visual Attention in DCNNs} Based on the concept of focus in human behavior, visual attention aims to highlight the important parts of an image. Literature suggests those important parts are chosen because they are significant to distinguish a specific image from others. The current research in visual attention aims to leverage those properties \cite{Attention_Branch_Network,Salient_Object_Detection,Multi_Channel_Attention_Selection_GAN,Pyramid_Feature_Attention,Channel_Attention_Based_Iterative_Residual_Learning,Squeeze_and_Attention_Networks,Unsupervised_Visual_Attention,Memory_Guided_Attention,Vision_Transformer_With_Deformable_Attention,Integration_Self_Attention_Convolution}. The highway network \cite{HighWay} introduced a basic --- yet effective --- gating mechanism that promotes the flow of information in deep neural networks. One can consider the ``transform gate" from the highway network as a type of attention. Based on ResNet backbone, SENet then presented channel attention using the squeeze and excitation architecture, ushering the start of a research wave aiming to improve the channel attention process. 

DANet \cite{DANet} integrated a position attention module with channel attention to model long-range contextual dependencies. NLNet \cite{NLNet} aggregated query-specific global context to each query position in the attention module. Building upon NLNet and SENet, GCNet \cite{GCNet} proposed the GC-block, which aims at capturing channel cross-talk and inter-dependencies while maintaining global context awareness. Triplet attention \cite{TripleAttention} explored an architecture that includes spatial and channel attention while maintaining computational efficiency. CBAM \cite{CBAM} applied global max-pooling instead of GAP in SENet as an alternative. GSoPNet \cite{GSoPNet} introduced global second-order pooling instead of GAP, which is more effective but computationally expensive. ECANet \cite{ECA-Net} remodeled the channel attention architecture to capture cross-channel interaction without unnecessary dimensionality reduction. 

FcaNet \cite{FcaNet} added a multi-spectral component to channel attention from a frequency analysis perspective. They explained the relationship between GAP and Discrete Cosine Transform's initial frequency and then used a selection of the remaining frequencies to extract channel information. Finally, WaveNet \cite{WaveNet} proposed to use Discrete Wavelet Transform to extract channel information. 

\subsection{Datasets for Visual Tasks} Behind every success in deep neural network tasks, there exists a dataset that was curated to guide those networks to generalize. ImageNet \cite{ImageNet} is considered the most popular image dataset mainly used for classification; it has more than $14$ million hand-annotated images. A very commonly used subset is ImageNet-1000 which has $1000$ classes, $1.3$ million images for training, and $50$ thousand images for validation. Another popular dataset is MS-COCO \cite{MS_COCO} consisting of approximately $118$ thousand annotated training images for object detection, key-points detection, and panoptic segmentation. It has $5$ thousand validation images. Places365 \cite{Places2} is a dataset for scene recognition, that consist of $1.8$ million training images from $365$ scene categories with $36$ thousand validation images. Finally, Birds \cite{Birds} is a medium size relatively easy classification dataset of different types of birds that are used to evaluate models on low-level features. The dataset consist of $450$ species, each with more than $150$ samples for training and five samples for validation. 

\newcommand*{\horzbar}{\rule[.5ex]{2.5ex}{0.5pt}}
\newcommand*{\vertbar}{\rule[-1ex]{0.5pt}{2.5ex}}

\newcommand{\FF}{\mathbf{F}}
\newcommand{\EE}{\mathbf{E}}
\newcommand{\rr}{\mathbb{R}}

\begin{figure*}[t]
    \centering
    % \includegraphics[width=0.9\textwidth]{figures/main.pdf}
    %\subfigure[FcaNet Module]{\includegraphics[width=0.93\textwidth]{figures/DCT_Attention.jpg}}
    \includegraphics[width=\textwidth]{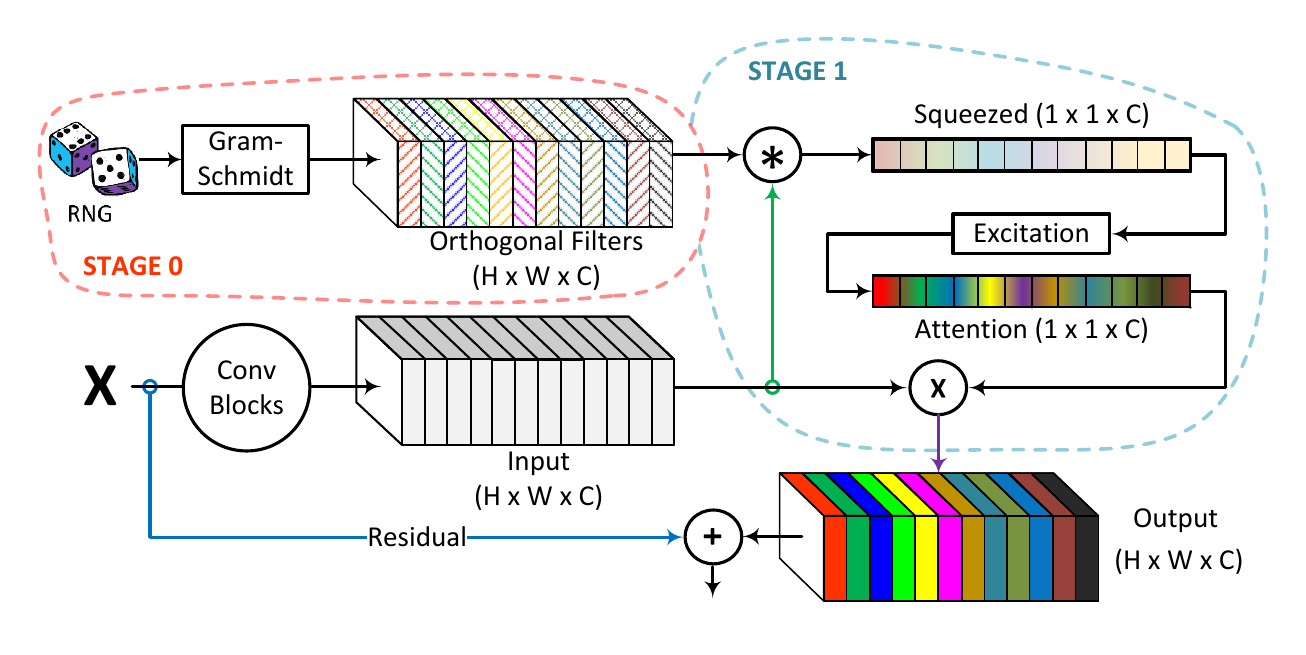}
    \caption{Illustration of orthogonal channel attention. Our method consists of two stages: Stage zero is initialization of random filters with sizes matching the feature layer sizes. We then use Gram-Schmidt process to make those filters orthonormal. Stage one utilizes those filters to extract the squeezed vector and use the excitation proposed by SENet to get the attention vector. By multiplying the attention vector with the input features, we calculate the weighted output features and add the residual.}
    \label{fig_main}
\end{figure*}

\section{Method}

In this section, we review the general formulation of channel attention mechanisms. With this formulation, we briefly review SENet and FcaNet. Based on these works, we introduce our method, Orthogonal Channel Attention, to be implemented in OrthoNet.

\subsection{Channel Attention}
\label{subsec_CA}
Considered a strongly influential attention mechanism, channel attention was first proposed in \cite{SENet} as an add-on module to be incorporated into any existing architecture. Its goal is to improve overall network performance with negligible computational cost. 

A \textit{channel attention block} is a computational unit, built to encapsulate information and highlight relevant features. Suppose $X \in \rr^C \times \rr^H \times \rr^W$ is a feature vector where $C$ is the number of channels with $H$ and $W$ being the height and width. The channel attention computes a vector, $A \in \rr^C$, which highlights the most relevant channels of $X$. The output of the module is $A \odot X$, calculated by 
\begin{equation}
    (A \odot X)_{c,h,w} = A_c X_{c,h,w}\,.
\end{equation}
Various researchers have proposed variants on Channel Attention that compute the attention vector, $A$, in different ways.

\subsection{Squeeze-and-Excitation (SENet)}
\label{subsec_SENet}
The Squeeze-and-Excitation method is broken into two stages: a \emph{squeeze} phase followed by an \emph{excitation} stage. The squeeze stage can be considered a lossy-compression method using GAP as shown below:
\begin{equation}\label{squeeze}
Z_c = \FF_{\text{se}}({X})_{c} = \frac{1}{H W}\sum_{i=1}^{H} \sum_{j=1}^{W} X_{c, i, j}\,.
\end{equation}
The excite step maps the compressed descriptor $Z$ to a set of channel weights. This is achieved by 
\begin{equation}\label{excitation}
\EE(Z) =  \sigma(W_2\delta(W_1 Z))\,,
\end{equation}
where $\sigma$ is the sigmoid function, $\delta$ is the ReLU activation function, and $W_1, W_2$ are (learnable) matrix weights. We can formulate channel attention in SENet in the following form:
\begin{equation}
\label{attentionSE}
    A(X) = \EE( \FF_{\text{se}}(X))\,.
\end{equation}
%cat GAP instead of downsample

\subsection{Frequency Channel Attention (FcaNet)}
\label{subsec_FcaNet}

One of the major weaknesses of the channel attention proposed in SENet \cite{SENet} was the use of only one squeeze method: GAP. The authors were motivated to use GAP to encourage representation of global information in the attention vector. FcaNet \cite{FcaNet} proposed an alternative squeeze method based on Discrete Cosine Transform. The DCT for an image of size $(H,W)$ with frequency component $(i,j)$ is given by
\begin{equation}
    T_{i,j}(X) = \sum_{h=1}^H \sum_{w=1}^W v_i(h, H) v_j(w, W) X_{h,w} \,,
\end{equation}
where
\begin{equation}
    v_{k}(a, A) = \cos\left(\frac{\pi k(a+1/2)}{A}\right) .
\end{equation}
They proved that GAP is proportional to the initial DCT frequency, $(0,0)$, and demonstrated that the remaining frequencies need to be represented in the attention vector. They claim that those missing frequencies contain essential information; however, they give no theoretical reasoning to explain why their choice frequencies was optimal. FcaNet attention vector can be computed as follows:

\begin{equation}
    A(X) = \EE(\FF_{\text{fca}}(X)), \label{attentionFCA}
\end{equation}
where 
\begin{equation}
    \FF_{\text{fca}}(X)_c = T_{I(c),J(c)}(X_c)
\end{equation}
for some choice of functions $I,J$ on the natural numbers. Based on experimental results, FcaNet chose to break the channel dimension into 16 blocks and chose $(I,J)$ to be constant on each block.

\input{algorithms/squeeze}
\subsection{Orthogonal Channel Attention (OrthoNet)}
\label{subsec_OrthoNet}

We notice that the DCTs used in FcaNet have a unique, essential, property: they are orthogonal \cite{DCT}. In this paper, we exploit the benefits of the orthogonal property for channel attention. Roughly, we start by randomly selecting filters of the appropriate dimension, $(C, H, W)$; we then apply Gram-Schmidt process to make those filters orthonormal. The full details for initializing the filters are described in algorithm \ref{algOrthoSqueeze}. Denoting these filters by $K \in \rr^C \times \rr^H \times \rr^W$, our squeeze process is given by
\begin{equation}\label{squeezeOrtho}
\FF_{\text{ortho}}(X)_c = \sum_{h=1}^H \sum_{w=1}^W K_{c, h, w} X_{c, h, w} \,.
\end{equation}
As in the other methods, we define our channel attention by 
\begin{equation}
    A(X) = \EE(\FF_{\text{ortho}}(X))
\end{equation}

%\color{black}

\input{tables/classification}

\section{Experimental Settings and Results}
We begin by describing the details of our experiments. We then report the effectiveness of our method on image classification, object detection, and instance segmentation tasks.

\subsection{Implementation Details}
\label{subsec_ImpDetails}

Following \cite{SENet,ECA-Net,FcaNet}, we add our proposed attention module to ResNet-34 to construct OrthoNet-34. Based on ResNet-50 and 101, we construct two versions of our network called OrthoNet and OrthoNet-MOD. They differ in the position of the attention module in the ResNet blocks. For further details refer to section \ref{MOD}.

\paragraph{General Specifications} We adopt the Nvidia APEX mixed precision training toolkit and Nvidia DALI library for training efficiency. Our models are implemented in PyTorch~\cite{PyTorch}, and are based on the code released by the authors of FcaNet~\cite{FcaNet}. The models were tested on two Nvidia Quadro RTX 8000 GPUs.

\subsubsection{Classification}

We utilize ImageNet~\cite{ImageNet}, Places365 \cite{Places2}, and Birds \cite{Birds} datasets to test and evaluate our method. To judge efficiency, we report the number of floating point operations per second (FLOPs) and the number of frames processed per second (FPS). To demonstrate method effectiveness, we report the top-1 and top-5 accuracies (T1, T5 acc). We use the same data augmentation and hyper-parameter settings found in \cite{FcaNet}. Specifically, we apply random horizontal flipping, random cropping, and random aspect ratio. The resulting images are of size $256\times256$. 

During training, the SGD optimizer is set with a momentum of $0.9$, the learning rate is $0.2$, the weight decay is $1e-4$, and the batch size is $256$. All models are trained for $100$ epochs using Cosine Annealing Warm Restarts learning schedule and label smoothing. At the beginning of every tenth epoch, the learning rate scales by $10\%$ of the previous learning rate. Doing so fosters convergence, as demonstrated in~\cite{FcaNet}.
\subsubsection{Object Detection and Segmentation}
We train and evaluate object detection and segmentation tasks using MS-COCO \cite{MS_COCO}. We report the average precision (AP) metric and its many variants.
\paragraph{FasterRCNN} We use FasterRCNN with OrthoNet-50 and 101 with one frozen stage and learnable BatchNorm. We use the same configuration used by FcaNet \cite{FcaNet} based on MMDetection toolbox \cite{MMDetection}. 

\paragraph{MaskRCNN} We use MaskRCNN with OrthoNet-50. We have one frozen stage and no learnable BatchNorm. We utilize the base configuration described in MMDetection toolbox \cite{MMDetection} of ResNet-50.

\paragraph{Training Configuration} We train for 12 epochs. The SGD optimizer is set with a momentum of $0.9$. We warm up the model in the first 500 iterations starting with a learning rate of $1e-4$ and growing by $0.001$ every 50 iterations. After, we set the learning rate to $0.01$ for the first 8 epochs. At epoch 9, the learning rate decays to $0.001$ then at epoch 12 it decays to $0.0001$. We evaluate both FcaNet and our method using this configuration. For more details, refer to section \ref{sec:limits}.

% The learning rate is $0.01$ with a weight decay of $1e-4$. Models are trained for $12$ epochs using a step learning scheduler with linear warm-up every 500 iterations and a warm up ratio of $0.001$ where the milestones to decay the learning rate is epoch 8 and 11.   

\subsection{Results}

\paragraph{Classification Results}

Accuracy results obtained on ImageNet are shown in \ref{classification}. The first noticeable feature is OrthoNet's superior performance when using ResNet-34 as the backbone. The result shown in the table is an average over five trials with a standard deviation of $0.12$. Even in the worst case, we achieved $74.97$ which is comparable to FcaNet. Our results on ResNet-50 and 101 are better than both FcaNet-LF and FcaNet-NAS. The result for OrthoNet-50 is a mean over four trials with standard deviation of $0.07$.

The results obtained on ResNet-50 validate our initial hypothesis. If we Consider a linear scale with SENet mapping to zero and FcaNet mapping to one, our method achieves a $0.93$. Since SENet has the least orthogonal filters (they are all the same), this demonstrates the importance of orthogonal filters. 

We believe that FcaNet slight improvement with ResNet-50 is because they choose particular filters that performed best on their experiments with ImageNet. To test this hypothesis, we evaluated our model on the Places365 and Birds classification datasets. Results are shown in \ref{classification_2}. We can see that our method outperforms FcaNet-TF on Birds by $0.18\%$ and on Places365 by $0.18\%$. These results imply that our method can generalize better to different datasets.  

\paragraph{Object Detection Results}

In addition to testing on ImageNet, Places365, and Birds datasets, we evaluate OrthoNet on MS COCO dataset to evaluate its performance on alternative tasks. We use OrthoNet and FPN \cite{FPN} as the backbone of Faster R-CNN and Mask R-CNN. 

As shown in Table \ref{detection}, when OrthoNet-MOD-50 is incorporated into the Faster-RCNN and Mask-RCNN frameworks performance surpasses that of FcaNet. We also achieve 10\% less computational cost.

\input{tables/classification_2}
\paragraph{Segmentation Results}

To further test our method, we evaluate OrthoNet-MOD-50 on instance segmentation task. As demonstrated in table \ref{segmentation}, our method outperforms FcaNet under the ResNet-50 basic configuration of Mask-RCNN framework. These results verify the effectiveness of our method.

%\color{black}

%\input{tables/classification}
\section{Experimental Settings and Results}
We begin by describing the details of our experiments. We then report the effectiveness of our method on image classification, object detection, and instance segmentation tasks.

\subsection{Implementation Details}
\label{subsec_ImpDetails}

Following \cite{SENet,ECA-Net,FcaNet}, we add our proposed attention module to ResNet-34 to construct OrthoNet-34. Based on ResNet-50 and 101, we construct two versions of our network called OrthoNet and OrthoNet-MOD. They differ in the position of the attention module in the ResNet blocks. For further details refer to section \ref{MOD}.

\subsubsection{General Specifications} We adopt the Nvidia APEX mixed precision training toolkit and Nvidia DALI library for training efficiency. Our models are implemented in PyTorch~\cite{PyTorch}, and are based on the code released by the authors of FcaNet~\cite{FcaNet}. The models were tested on two Nvidia Quadro RTX 8000 GPUs.

\subsubsection{Classification}

We utilize ImageNet~\cite{ImageNet}, Places365 \cite{Places2}, and Birds \cite{Birds} datasets to test and evaluate our method. To judge efficiency, we report the number of floating point operations per second (FLOPs) and the number of frames processed per second (FPS). To demonstrate method effectiveness, we report the top-1 and top-5 accuracies (T1, T5 acc). We use the same data augmentation and hyper-parameter settings found in \cite{FcaNet}. Specifically, we apply random horizontal flipping, random cropping, and random aspect ratio. The resulting images are of size $256\times256$. 

During training, the SGD optimizer is set with a momentum of $0.9$, the learning rate is $0.2$, the weight decay is $1e-4$, and the batch size is $256$. All models are trained for $100$ epochs using Cosine Annealing Warm Restarts learning schedule and label smoothing. At the beginning of every tenth epoch, the learning rate scales by $10\%$ of the previous learning rate. Doing so fosters convergence, as demonstrated in~\cite{FcaNet}.
\input{tables/object_detection}
\subsubsection{Object Detection and Segmentation}
We train and evaluate object detection and segmentation tasks using MS-COCO \cite{MS_COCO}. We report the average precision (AP) metric and its many variants.
\paragraph{FasterRCNN} We use FasterRCNN with OrthoNet-50 and 101 with one frozen stage and learnable BatchNorm. We use the same configuration used by FcaNet \cite{FcaNet} based on MMDetection toolbox \cite{MMDetection}. 

\paragraph{MaskRCNN} We use MaskRCNN with OrthoNet-50. We have one frozen stage and no learnable BatchNorm. We utilize the base configuration described in MMDetection toolbox \cite{MMDetection} of ResNet-50.

\subsubsection{Training Configuration} We train for 12 epochs. The SGD optimizer is set with a momentum of $0.9$. We warm up the model in the first 500 iterations starting with a learning rate of $1e-4$ and growing by $0.001$ every 50 iterations. After, we set the learning rate to $0.01$ for the first 8 epochs. At epoch 9, the learning rate decays to $0.001$ then at epoch 12 it decays to $0.0001$. We evaluate both FcaNet and our method using this configuration. For more details, refer to section \ref{sec:limits}.

% The learning rate is $0.01$ with a weight decay of $1e-4$. Models are trained for $12$ epochs using a step learning scheduler with linear warm-up every 500 iterations and a warm up ratio of $0.001$ where the milestones to decay the learning rate is epoch 8 and 11.   

\subsection{Results}

\paragraph{Classification Results}
Accuracy results obtained on ImageNet are shown in \ref{classification}. The first noticeable feature is OrthoNet's superior performance when using ResNet-34 as the backbone. The result shown in the table is an average over five trials with a standard deviation of $0.12$. Even in the worst case, we achieved $74.97$ which is comparable to FcaNet. Our results on ResNet-50 and 101 are better than both FcaNet-LF and FcaNet-NAS. The result for OrthoNet-50 is a mean over four trials with standard deviation of $0.07$.

The results obtained on ResNet-50 validate our initial hypothesis. If we Consider a linear scale with SENet mapping to zero and FcaNet mapping to one, our method achieves a $0.93$. Since SENet has the least orthogonal filters (they are all the same), this demonstrates the importance of orthogonal filters. 

We believe that FcaNet slight improvement with ResNet-50 is because they choose particular filters that performed best on their experiments with ImageNet. To test this hypothesis, we evaluated our model on the Places365 and Birds classification datasets. Results are shown in \ref{classification_2}. We can see that our method outperforms FcaNet-TF on Birds by $0.18\%$ and on Places365 by $0.18\%$. These results imply that our method can generalize better to different datasets.  
\input{tables/instance_segmentation}
\paragraph{Object Detection Results}

In addition to testing on ImageNet, Places365, and Birds datasets, we evaluate OrthoNet on MS COCO dataset to evaluate its performance on alternative tasks. We use OrthoNet and FPN \cite{FPN} as the backbone of Faster R-CNN and Mask R-CNN. 

As shown in Table \ref{detection}, when OrthoNet-MOD-50 is incorporated into the Faster-RCNN and Mask-RCNN frameworks performance surpasses that of FcaNet. We also achieve 10\% less computational cost.

\paragraph{Segmentation Results}

To further test our method, we evaluate OrthoNet-MOD-50 on instance segmentation task. As demonstrated in table \ref{segmentation}, our method outperforms FcaNet under the ResNet-50 basic configuration of Mask-RCNN framework. These results verify the effectiveness of our method.

\section{Discussion}

We begin with a possible explanation for the success of OrthoNet. We then detail our explorations into variants of our architecture and conclude by discussing implementation and limitations.

\subsection{Potential Mechanism for Our Success}
While FcaNet believes that frequency choice for the discrete cosine transform is the primary factor for a successful attention mechanism, our results imply that the key driving force behind successful attention mechanisms is distinct (orthogonal) attention filters. In brief, we believe that the success of FcaNet is mostly due to the orthogonality of the DCT kernels. 

Recall that convolutional neural networks contain built-in redundancy \cite{Redundancy1,Redundancy2,FcaNet} (i.e., hidden features with strongly correlated channel slices). In a standard SENet, the squeeze method will extract the same information from these redundant channels. In fact, by appropriately permuting the learnable parameters of a SENet, one can construct another network where the only difference is the order of the channels in the intermediate features. (Such a trick cannot be applied to FcaNet and OrthoNet due to the presence of constant, orthogonal squeeze filters.) The existence of this permutation method implies that SENet cannot, a priori, prescribe meaning to its channels. 

When the filters are orthogonal, the information they extract comes from orthogonal subspaces of the feature space ($\mathbb{R}^{H} \times \mathbb{R}^{W}$). Hence, they focus distinct characteristics. Since the gradient information flows backward through the network, the convolutional kernels preceding the squeeze can adapt to their unique mapping. By doing so, the network can extract a richer representation for every feature map, which the excitation can then build upon.

To further support our ideas, we train OrthoNet-34 and 50 using random filters --- we omit the Gram-Schmidt step --- and obtain accuracies of $74.63$ and $76.77$ respectively. We can clearly see the effect of orthogonal filters on improving the validation accuracy on both networks compared to random filter and GAP (Table \ref{classification}).

% Their flagship network FcaNet-TF achieved highest accuracy when tailored for ImageNet. Looking at the other variants of the network, we can see that by switching the selection of the frequency components, the network accuracy drops.   

% we believe that the rank of the frequency choice is not the primary reason for successful attention rather, the orthogonality of those frequency components are suitable for extracting the best compressed vector tailored for ImageNet dataset. 

%\subsection{Architecture Choices}

\subsection{Attention Module Location}
\label{MOD}
\begin{figure}
\centering
\begin{minipage}{.5\textwidth}
  \centering
  \includegraphics[width=\linewidth]{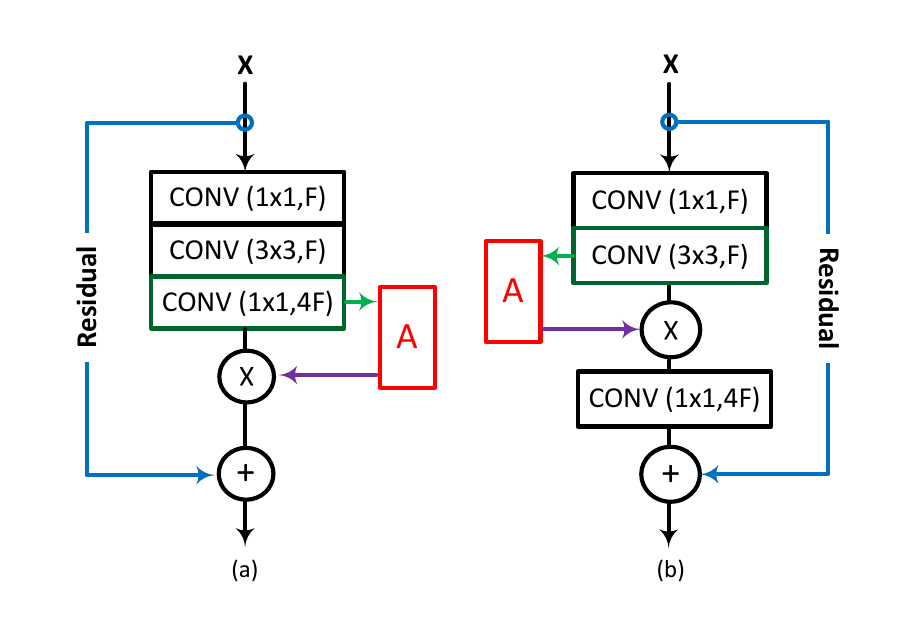}
  \caption{(a) OrthoNet block vs (b) OrthoNet-MOD block }
  \label{fig:OGvMOD}
\end{minipage}
\end{figure}

The overall structure of a ResNet-50 or 101 bottleneck block is shown in figure \ref{fig:OGvMOD}. To construct OrthoNet, we follow the standard procedure placing the attention module following the $1\times1$ convolution. Motivated by the below reasoning, we moved the attention in those networks to follow the $3 \times 3$ convolution. The resulting network is OrthoNet-MOD.

In OrthoNet-34, the attention is placed on the second $3 \times 3$ convolution in each block. We now recall the structure of ResNet 50 and 101: both networks contains many blocks consisting of a $1 \times 1$ convolution followed by a $3 \times 3$ convolution then another $1 \times 1$ convolution (with batch-norms and activations placed between). Since the $1 \times 1$ convolutions can only consider inter-channel relationships and lack the ability to capture spatial information, they are mainly used for feature refinement and for channel resizing. Combined with an activation, a $1 \times 1$ convolution can be considered as a Convolutional MLP \cite{NIN}. Since the $3 \times 3$ convolutions are the only modules that consider spatial-relationships, they must extract rich spacial information if the network hopes to achieve high accuracies. These motivations and the success of OrthoNet-34 lead to the construction of OrthoNet-MOD.

This modification yields many benefits. First, we reduce the number of parameters used for the attention module by only creating filters of the same size as we did in OrthoNet-34. Exact parameter counts can be found in Table \ref{classification}. Second, we improve accuracies over the standard OrthoNet. Third, we place the attention at a more meaningful location, the $3 \times 3$ convolution, where features are richer in spacial information.

\subsection{Cross-talk Effect on Orthogonal Filters}
Our squeeze method can be considered as a convolution with kernel size equal to the feature spacial dimensions and groups equal to the number of channels --- group size equal to one.
Since increasing group size allows inter-communication between channels; one might hypothesize that doing so could allow the squeeze step to extract richer representations of the input feature. To investigate such a hypothesis on OrthoNet, we conduct experiments using OrthoNet-34 with different grouping sizes. The results are recorded in Table \ref{Grouping_Effect}. The results indicate that the group size is inconsequential; however, due to time and computational constraints, we were only able to run each experiment once, and more experiments need to be conducted.
\input{tables/Grouping_Effect}
\subsection{Fine-Tuning channel attention filters}

In this section, we experiment with allowing the orthogonal filters to learn and fine-tune. We implement multiple different learning methods. First, we train OrthoNet-34 on constant orthogonal filters, then introduce learning the filters in the last twenty epochs. We call this method FineTuned-20. Second, we implement FineTuned-40, where we learn during each epoch that's divisible by five and also for the last twenty epochs --- 40 epochs in total. Third, we implement learning the first 30 epochs --- FineTuned-30 --- then we disable learning the filters and continue training for the remaining 70 epochs on OrthoNet-50. Experiments demonstrate that learning the filters for OrthoNet-MOD-50 doesn't improve the overall validation accuracy. As for OrthoNet-34, we observe a more consistent accuracies for different learning methods. Results are shown in table \ref{Learning_Effect}. Further investigation is needed on the method of training and the potential inclusion of an attention-specific loss function for optimal learning. %
\input{tables/Learning_Effect}

\subsection{Ease of Integration}

Similar to SENet and FcaNet, our module can easily be integrated to any existing convolutional networks. The major distinction between SENet, FcaNet and our method is the adoption of different channel compression methods. As discussed earlier, our method can be described as a convolution as shown in \ref{algOrthoSqueeze} and can be implemented with a single line of code.

\subsection{Limitations} 
\label{sec:limits}
Random filters have a natural limitation. In upper layers, where $HW>C$, it is impossible to have filters that comprise a full basis (i.e., you must choose $C$ filters from the $HW$ total). Although we did not observe it, a poor choice of filters may exist. Their unlearnable nature makes this a persistent factor throughout training. This could lead to a failure case. However; in the lower layers, we are able --- and do --- create a complete basis for $\mathbb{R}^{HW}$ which eliminates this issue in those layers.

Although we reported the means for OrthoNet-34 over five trials and OrthoNet-50 over four trials; due to limitations in time and computational capabilities, we only ran most other experiments once. Due to the limited number of runs, we could not report the standard deviation. However, OrthoNet’s constant higher accuracy on a variety of tasks and datasets suggest the robustness of our network. 

For the object detection and segmentation with MaskRCNN, we utilize the base configuration provided by MMDetection \cite{MMDetection}. The results reported by the base configuration are different from those of their configuration. We were unable to use the configuration used by FcaNet \cite{FcaNet} because of technical difficulties. 

We also used the results of the experiments conducted by FcaNet for previous methods (FasterRCNN and ImageNet); however, we conducted our experiments under the same configuration.

\section{Conclusion}

In this work, we introduced a variant of SENet which utilizes orthogonal squeeze filters to  create an effective channel attention module that can be integrated into any existing network. To evaluate its performance, we constructed OrthoNet and demonstrated state-of-the-art performance on Birds, Places365, and COCO with competitive or superior performance on ImageNet. By comparing OrthoNet to state-of-the-art, we've found that the key driving force to a successful attention mechanism is orthogonal attention filters. Orthogonal filters allow for maximum information extraction from any correlated features and allow the network to prescribe meaning to each channel yielding a more effective attention squeeze. 

Our future works include further investigating learnable orthogonal filters, implementing metrics for evaluation of attention mechanisms, and further pruning our attention module to lower the computational cost and improve our method accuracy. We're also searching for a theoretical framework to explain the channel attention phenomenon. 

{\small
\bibliographystyle{ieee_fullname}
\bibliography{egbib}
}

\end{document}

%% file: algorithms/squeeze.tex
\begin{algorithm}

\caption{Orthogonal Channel Attention Initialization (Stage Zero)}
\label{algOrthoSqueeze}
\SetAlgoLined
\KwIn{Input Feature Dimension: $(C, H, W)$.}
\KwOut{Kernel  $K \in \rr^C \times \rr^H \times \rr^W$.}
\eIf{$HW < C$}
{
Calculate $n = \text{floor}(C / (HW))$\\
Initialize $\mathbf{L}$ as empty list\\
\For{$i \in  \{0 \dots n \}$}
{
    Initialize $HW$ random filters $F_j \in \rr^{HW}$\\
    Run Gram-Schmidt process on $\{F_j\}_{j=1}^{HW}$ to get an orthogonal set $\{F'\}_{j=1}^{HW}$.\\
    Append $\{F'\}_{j=1}^{HW}$ to List $L$.\\
}
Concatenate filters in list $L$ to get kernel $K \in \rr^C \times \rr^H \times \rr^W$.
}
{    
    Initialize $C$ random filters $F_j \in \rr^{HW}$\\
    Run Gram-Schmidt process on $\{F_j\}_{j=1}^{C}$ to get an orthogonal set $\{F'\}_{j=1}^{C}$.\\
    Concatenate filters $\{F'\}_{j=1}^C$ to get kernel $K \in \rr^C \times \rr^H \times \rr^W$.
}

\end{algorithm}

%% file: tables/classification.tex
%ecanet34     21797720 Params  3677019200.0 Mac
%freqnet34    21954856 Params  3678545132.0 Mac
%senet34      21954856 Params  3677165292.0 Mac

%ecanet50    25557080 Params  4127488768.0 Mac
%freqnet50   28071976 Params  4135478704.0 Mac
%senet50     28071976 Params  4129959344.0 Mac
% 5.264 MMac
%freqnet101   49292328 Params  7872106992.0 Mac
%senet101     49292328 Params  7863175664.0 Mac
%ecanet101    44549259 Params  7858528000.0 Mac
% 8.518 MMac
%freqnet152    66770984 Params  11613560240.0 Mac
%senet152      66770984 Params  11600414128.0 Mac
%ecanet152     60192958 Params  11593976576.0 Mac
% 12.537 MMac

\begin{table*}[htbp]
	\centering
	\caption{Results of the image the classification task on ImageNet over different methods. }
	\label{classification}
	\begin{threeparttable}                                        
    \begin{tabular}{lcccccccc}
	\toprule
	Method & Years & Backbone & Parameters & FLOPS & Train FPS & Test FPS & T1 acc & T5 acc \\
	\hline
	ResNet~\cite{ResNet}    & CVPR16 & \multirow{8}{*}{ResNet-34} & 21.80 M & 3.68 G & 2898 & 3840 & 74.58 & 92.05 \\
	SENet~\cite{SENet}      & CVPR18 &                            & 21.95 M & 3.68 G & 2729 & 3489 & 74.83 & 92.23 \\
	ECANet~\cite{ECA-Net}   & CVPR20 &                            & 21.80 M & 3.68 G & 2703 & 3682 & 74.65 & 92.21\\ 
	FcaNet-LF~\cite{FcaNet} & ICCV21    &                         & 21.95 M & 3.68 G & 2717 & 3356 & 74.95 & 92.16\\
	FcaNet-TS~\cite{FcaNet} & ICCV21    &                         & 21.95 M & 3.68 G & 2717 & 3356 & 75.02 & 92.07\\
	FcaNet-NAS~\cite{FcaNet}& ICCV21    &                         & 21.95 M & 3.68 G & 2717 & 3356 & 74.97 & 92.34\\
	WaveNet-C~\cite{WaveNet}& Bigdata22 &                         & 21.95 M & 3.68 G & 2717 & 3356 & 75.06 & 92.376\\
	\rowcolor{tb_bg_color}
	OrthoNet\tnote{+}       & CVPR22    &                         & 21.95 M & 3.68 G & 2717 & 3356 & \textbf{75.22} & \textbf{92.5}\\
	% & Training & Inference 
	\hline
	ResNet~\cite{ResNet}    & CVPR16    & \multirow{12}{*}{ResNet-50} & 25.56 M & 4.12 G & 1644 & 3622 & 77.27 & 93.52\\
	SENet~\cite{SENet}      & CVPR18    &                             & 28.07 M & 4.13 G & 1457 & 3417 & 77.86 & 93.87\\
	CBAM~\cite{CBAM}        & ECCV18    &                             & 28.07 M & 4.14 G & 1132 & 3319 & 78.24 & 93.81\\
	GSoPNet1\tnote{*}~\cite{GSoPNet} & CVPR19 &                       & 28.29 M & 6.41 G & 1095 & 3029 & 79.01 & 94.35\\
	GCNet~\cite{GCNet}      & ICCVW19   &                             & 28.11 M & 4.13 G & 1477 & 3315 & 77.70 & 93.66\\ 
	AANet~\cite{AANet}      & ICCV19    &                             & 25.80 M & 4.15 G & -    & -    & 77.70 & 93.80\\
	ECANet~\cite{ECA-Net}   & CVPR20    &                             & 25.56 M & 4.13 G & 1468 & 3435 & 77.99 & 93.85\\ 
	FcaNet-LF~\cite{FcaNet}& ICCV21     &                             & 28.07 M & 4.13 G & 1430 & 3331 & 78.43 & 94.15\\
	FcaNet-TS~\cite{FcaNet}& ICCV21     &                             & 28.07 M & 4.13 G & 1430 & 3331 & \textbf{78.57} & 94.10\\
	FcaNet-NAS~\cite{FcaNet}& ICCV21    &                             & 28.07 M & 4.13 G &  & 3331 & 78.46 & 94.09\\
	\rowcolor{tb_bg_color}
	OrthoNet\tnote{+}       & CVPR22   &                             & 28.07 M & 4.13 G & 1430 & 3331 & 78.47 & 94.12\\
	\rowcolor{tb_bg_color}
	OrthoNet-MOD\tnote{+\$}   & CVPR22   &                             & 25.71 M & 4.12 G & 1430 & 3331 & 78.52 & \textbf{94.17}\\
	\hline
	ResNet~\cite{ResNet}    & CVPR16    & \multirow{10}{*}{ResNet-101}& 44.55 M & 7.85 G & 816  & 3187 & 78.72 & 94.30\\
	SENet~\cite{SENet}      & CVPR18    &                             & 49.29 M & 7.86 G & 716  & 2944 & 79.19 & 94.50\\
	CBAM~\cite{CBAM}        & ECCV18    &                             & 49.30 M & 7.88 G & 589  & 2492 & 78.49 & 94.31\\
	AANet~\cite{AANet}      & ICCV19    &                             & 45.40 M & 8.05 G &  -   &   -  & 78.70 & 94.40\\
	ECANet~\cite{ECA-Net}   & CVPR20    &                             & 44.55 M & 7.86 G & 721  & 3000 & 79.09 & 94.38\\ 
	FcaNet-LF\cite{FcaNet}  & ICCV21    &                             & 49.29 M & 7.86 G & 705  & 2936 & 79.46 & 94.60\\
	FcaNet-TS\cite{FcaNet}  & ICCV21    &                             & 49.29 M & 7.86 G & 705  & 2936 & 79.63 & 94.63\\
	FcaNet-NAS\cite{FcaNet} & ICCV21    &                             & 49.29 M & 7.86 G & 705  & 2936 & 79.53 & 94.64\\
	\rowcolor{tb_bg_color}
	OrthoNet\tnote{+}       & CVPR22    &                             & 49.29 M & 7.86 G & 705  & 2936 & 79.61 & \textbf{94.73}\\
	\rowcolor{tb_bg_color}
	OrthoNet-MOD\tnote{+\$}   & CVPR22    &                             & 44.84 M & 7.85 G & 705 & 2936 & \textbf{79.69} & 94.61\\
	\bottomrule
	\end{tabular}
	\begin{tablenotes}
	    \item \footnotesize \centering
		* High computational cost. + Randomly initialized. \$ Reduced number of parameters compared to SOTA.
	\end{tablenotes}
\end{threeparttable}

\end{table*}

%% file: tables/classification_2.tex
\begin{table}[htbp]
	\centering
	\caption{Results of Places365 dataset and  Birds dataset. Our method achieves superior performance compared to FCANet. Both methods are trained using ResNet-50 backbone.}
	\label{classification_2}
    \begin{tabular}{lccc}
    	\toprule
            
    	Method & Dataset & T1 acc & T5 acc \\
    	\hline
    	FcaNet-TS~\cite{FcaNet}& Places365 \cite{Places2} &  56.15          & \textbf{86.19}\\
    	\rowcolor{tb_bg_color}
    	OrthoNet-MOD           & Places365 \cite{Places2} & \textbf{56.33} 
    	&86.18\\
    	\hline
		FcaNet-TS~\cite{FcaNet}& Birds \cite{Birds} &  97.60          & 99.47\\
    	\rowcolor{tb_bg_color}
    	OrthoNet-MOD           & Birds \cite{Birds} & \textbf{97.78} & \textbf{99.64}\\
    	\bottomrule
    \end{tabular}
\end{table}

%% file: tables/object_detection.tex
\begin{table*}[htbp]
	\centering
	\caption{Results of the object detection task on COCO val 2017 over different methods.}
	\label{detection}
	\begin{tabular}{lccccccccc}
		\toprule
		Method  & Detector & Parameters & FLOPs & AP & $AP_{50}$ & $AP_{75}$ & $AP_{S}$ & $AP_{M}$ &$AP_{L}$\\
		% & Training & Inference 
		\hline
		ResNet-50 & \multirow{5}{*}{Faster-RCNN} & 41.53 M  & 215.51 G &  36.4   & 58.2 & 39.2 & 21.8 & 40.0 & 46.2\\
		SENet &  & 44.02 M & 215.63 G & 37.7 & 60.1 & 40.9 & 22.9 & 41.9 & 48.2\\
		ECANet & & 41.53 M & 215.63 G & 38.0 & 60.6 & 40.9 & 23.4 & 42.1 & 48.0\\
		FcaNet-TS  &  & 44.02 M & 215.63 G & 39.0 & 61.1 & 42.3 & 23.7 & 42.8 & 49.6\\
		\rowcolor{tb_bg_color}
		OrthoNet-MOD  &  & 41.68 M & 215.63 G & \textbf{39.1} & 60.2 & 42.5 & 23.4 & 43.1 & 49.7\\
		\hline
        ResNet101 & \multirow{5}{*}{Faster-RCNN} & 60.52 M & 295.39 G &  38.7 & 60.6 &41.9 &22.7 &43.2 &50.4\\
        SENet &  & 65.24 M & 295.58 G & 39.6 &62.0 &43.1 &23.7 &44.0 &51.4\\
        ECANet & & 60.52 M & 295.58 G & 40.3 & 62.9 & 44.0 & 24.5 & 44.7 &51.3\\
        FcaNet-TS  &  & 65.24 M & 295.58 G & \textbf{41.2} & 63.3 & 44.6 & 23.8 & 45.2 & 53.1\\
        \rowcolor{tb_bg_color}
        OrthoNet-MOD  &  & 60.84 M & 295.58 G & 40.8 & 61.6 & 44.9 & 24.7 & 44.8 & 51.8\\
		\hline
		FcaNet-TS-50$^*$    &  & 46.66 M & 261.93 G & 37.3 & 57.5 & 40.5 & 22.1 & 40.4 & 48.2\\
		\rowcolor{tb_bg_color}
		OrthoNet-MOD$^*$  & \multirow{-2}{*}{Mask-RCNN} & 44.53 M & 261.93 G & \textbf{37.8} & 57.9 & 41.1 & 22.5 & 41.3 & 48.2\\
		\bottomrule
	\end{tabular}
	\begin{tablenotes}
	    \item \footnotesize \centering
		* Base configuration used for training. For more details refer to section \ref{subsec_ImpDetails} and section \ref{sec:limits}.
	\end{tablenotes}
\end{table*}

%% file: tables/instance_segmentation.tex
% \begin{table}[h]
% 	\centering
% 	\caption{Instance segmentation results of different methods using Mask R-CNN on COCO val 2017.}
% 	\label{segmentation}
% 	\setlength{\tabcolsep}{0.7mm}{\begin{tabular}{l|c|c|c|c|c|c}
% 		\toprule
% 		Method & AP & $AP_{50}$ & $AP_{75}$ & $AP_{S}$ & $AP_{M}$ &$AP_{L}$\\
% 		% & Training & Inference 
% 		\hline
% 		ResNet-50\ \ \   & 34.1 & 55.5 & 36.2 & 16.1 & 36.7 & 50.0\\
% 		SENet\ \ \  & 35.4 & 57.4 & 37.8 & 17.1 & 38.6 & \textbf{51.8}\\
% 		GCNet\ \ \  & 35.7 & 58.4 & 37.6 & N/A & N/A & N/A\\
% 		ECANet\ \ \  & 35.6 & 58.1 & 37.7 & 17.6 & 39.0 & \textbf{51.8}\\
% 		FcaNet (Ours)\ \ \  & \textbf{36.2} & \textbf{58.6} & \textbf{38.1} & \textbf{20.9} & \textbf{39.7} & 48.7\\
% 		\bottomrule

% 	\end{tabular}}
% \end{table}

\begin{table}[h]
	\centering
	\caption{Results of the instance segmentation task on COCO val 2017 over different methods using Mask R-CNN.}
	\label{segmentation}
	\begin{tabular}{lccccc}
		\toprule
		Method    & AP            & AP$_{50}$ & AP$_{75}$ & AP$_{S}$ & AP$_{M}$ \\
		% & Training & Inference 
		\hline
		FcaNet-TS-50 &         34.0  & 54.6      & 36.1      &    18.5  & 37.1      \\
		
        \rowcolor{tb_bg_color}
		OrthoNet-MOD & \textbf{34.4} & 55.0      & 36.8      &    19.0  & 37.5      \\
		\bottomrule
	\end{tabular}
	\begin{tablenotes}
	    \item \footnotesize
		Base configuration used for training. For more details refer to 
		\item \footnotesize
		section \ref{subsec_ImpDetails} and section \ref{sec:limits}.
	\end{tablenotes}
\end{table}

% 0.340 0.546 0.361 0.185 0.371 0.455
% 0.344 0.550 0.368 0.190 0.375 0.460

%% file: tables/Grouping_Effect.tex
% \begin{table}[htbp]
% 	\centering
% 	\caption{Effect of grouping Orthogonal channel attention filters using ResNet-34 and ResNet-50 trained on ImageNet.}
% 	\label{Grouping_Effect}
%     \begin{tabular}{lcc}
%     	\toprule
%     	Method                       &  Group Size  & Top-1 acc \\
%     	\hline
%     	\multirow{3}{*}{OrthoNet-34} &     1        &  75.13    \\
%     	                             &     4        &  75.20    \\
%     	                             & $H\times W$  &  75.18    \\
%     	\hline
%     	\multirow{3}{*}{OrthoNet-50} &     1        &  78.34    \\
%     	                             &     4        &  78.39    \\
%     	                             & $H\times W$  &  78.30    \\
%     	\bottomrule
%     \end{tabular}
% \end{table}

\begin{table}[htbp]
	\centering
	\caption{Effect of Grouping.}
	\label{Grouping_Effect}
    \begin{tabular}{lccccccc}
    	
    	Method                      & & & &Group Size  & & &   Top-1 acc\\
    	\toprule
		\multirow{3}{*}{OrthoNet-34}& & & &1           & & &   75.13\\
		                            & & & &4           & & &   75.20\\
            		                & & & &$H \times W$& & &   75.18\\
    	\bottomrule
    \end{tabular}
\end{table}

%% file: tables/Learning_Effect.tex
\begin{table}[htbp]
	\centering
	\caption{Effect of Squeeze Filter Learning.}
	\label{Learning_Effect}
    \begin{tabular}{lcc}
    	Backbone        & Learning Method & Top-1 acc \\
    	\toprule
		OrthoNet-34     & FineTuned-20    &   75.20  \\
		OrthoNet-34     & FineTuned-40    &   75.24  \\
        \hline
        OrthoNet-50     & FineTuned-40    &   78.50  \\
		OrthoNet-MOD-50 & FineTuned-40    &   78.30  \\
		OrthoNet-MOD-50 & FineTuned-30    &   78.36  \\
    	\bottomrule
    \end{tabular}
\end{table}